\documentclass[conference]{IEEEtran}
\IEEEoverridecommandlockouts

\usepackage{cite}
\usepackage{amsmath,amssymb,amsfonts}
\usepackage{algorithmic}
\usepackage{graphicx}
\usepackage{textcomp}
\usepackage{xcolor}
\usepackage{booktabs}
\usepackage{listings}
\usepackage{hyperref}
\usepackage{float}

\def\BibTeX{{\rm B\kern-.05em{\sc i\kern-.025em b}\kern-.08em
    T\kern-.1667em\lower.7ex\hbox{E}\kern-.125emX}}

\begin{document}

\title{MeddeX: A Retrieval-Augmented LLaMA-3 System with Gemini Vision for Evidence-Backed Medical Report Analysis}

\author{
  \IEEEauthorblockN{Rajdeep Pandhare\textsuperscript{1}, Rohit Kundaikar\textsuperscript{2}, Yakub Kojar\textsuperscript{3}, Shreyansh Tiwari\textsuperscript{4}}
  \IEEEauthorblockA{
    Department of Artificial Intelligence \& Data Science\\
    Rajiv Gandhi Institute of Technology, Mumbai, India\\
    \textit{Under the Guidance of}\\
    \textbf{Prof. Naresh Shende}\\
    Department of Artificial Intelligence \& Data Science\\
    Rajiv Gandhi Institute of Technology, Mumbai, India
  }
}

\maketitle

\begin{abstract}
Modern healthcare generates increasingly complex medical reports that challenge clinical professionals in terms of time and interpretive burden. Manual review processes are slow, error-prone, and difficult to scale. This paper presents \textbf{MeddeX}, a Medical Report Analyzer that combines a QLoRA fine-tuned LLaMA-3 (8B) model with Retrieval-Augmented Generation (RAG) and a multimodal Gemini Vision layer to deliver accurate, concise, and evidence-backed summaries of medical documents. The system handles text-native PDFs, scanned image-based PDFs, and direct radiological images through an intelligent hybrid extraction pipeline that includes a novel garbage-text detection mechanism for corrupted and UTF-16-encoded documents. Beyond single-report Q\&A, MeddeX introduces report-scoped retrieval for precise contextual grounding, an automated longitudinal report comparison engine, and a self-healing backfill pipeline that lazily processes legacy documents on demand. A FastAPI backend, Supabase vector store, and Next.js frontend complete the end-to-end architecture. Evaluation against the base LLaMA3-Med42-8B model demonstrates consistent improvements across ROUGE-1 (+9.2\%), ROUGE-2 (+10.0\%), ROUGE-L (+11.8\%), BLEU (+71.4\%), and BERTScore (+0.8\%), validating the benefit of domain-specific fine-tuning. MeddeX represents a practical and robust step toward AI-assisted clinical decision support.
\end{abstract}


\section{Introduction}

Healthcare professionals routinely contend with lengthy, complex medical reports encompassing diagnostic results, patient histories, lab values, and clinical recommendations. The cognitive load imposed by manual review slows clinical workflows and increases the risk of oversight-related errors. With the exponential growth of digital health records, intelligent systems capable of rapid, reliable medical document interpretation have become a clinical necessity.

Recent advances in Large Language Models (LLMs) and Retrieval-Augmented Generation (RAG) have opened new possibilities for automated medical report analysis. However, existing solutions suffer from well-documented limitations: poor handling of long-context documents, absence of verifiable evidence linking, inadequate adaptation to specialized medical terminology, and limited support for multimodal inputs such as scanned reports or radiological images.

This paper presents \textbf{MeddeX}, a system that directly addresses these limitations. MeddeX integrates:
\begin{itemize}
    \item A \textbf{QLoRA fine-tuned LLaMA-3 (8B)} model (based on LLaMA3-Med42-8B) for domain-adapted generation.
    \item A \textbf{RAG pipeline} with report-scoped dense retrieval to ground outputs in specific source evidence.
    \item A \textbf{Gemini Vision} multimodal layer for OCR on scanned PDFs, UTF-16-encoded documents, and direct image inputs.
    \item A \textbf{garbage-text detection} module that identifies and re-routes corrupted or hex-encoded page extractions.
    \item A \textbf{longitudinal comparison engine} (\texttt{/compare\_reports}) enabling AI-driven analysis of change across multiple reports over time.
    \item A \textbf{lazy backfill pipeline} that transparently processes legacy documents uploaded before the chunking system was in place.
    \item A \textbf{FastAPI} backend with \textbf{Supabase} vector and document storage, and a \textbf{Next.js} frontend for clinician-friendly interaction.
\end{itemize}

The remainder of this paper is organized as follows. Section~\ref{sec:lit} reviews related work and identifies research gaps. Section~\ref{sec:system} details the proposed architecture. Section~\ref{sec:impl} describes key implementation components. Section~\ref{sec:eval} presents experimental results. Section~\ref{sec:conc} concludes and outlines future directions.

\section{Literature Review}
\label{sec:lit}

\subsection{Medical NLP and Summarization}

Natural Language Processing for clinical text has been studied extensively. Early systems relied on rule-based extraction and keyword matching, which lacked generalizability. The introduction of transformer architectures, beginning with BERT \cite{devlin2019bert}, enabled contextual understanding of medical language. Subsequent models such as BioBERT \cite{lee2020biobert}, ClinicalBERT, and PubMedBERT demonstrated that domain pretraining significantly improves downstream clinical NLP tasks including named entity recognition, relation extraction, and question answering.

Summarization-specific models including BART \cite{lewis2020bart} and Longformer \cite{beltagy2020longformer} have been applied to medical documents, with Longformer's sliding-window attention providing a partial solution to the long-context problem. However, these models generate summaries without grounding claims in verifiable source passages, limiting clinician trust.

\subsection{Retrieval-Augmented Generation}

RAG, introduced by Lewis et al.\ \cite{lewis2020rag}, combines a dense retriever with a seq2seq generator to produce factually grounded responses. The formulation conditions generation on retrieved passages:
\begin{equation}
P(y \mid x) = \sum_{z \in \mathcal{D}} P(y \mid x, z;\, \theta)\, P(z \mid x;\, \phi)
\label{eq:rag}
\end{equation}
where $x$ is the input query, $z$ represents retrieved document chunks, and $\phi$, $\theta$ are the retriever and generator parameters respectively. This joint framework enables outputs to be traced to source evidence, a critical requirement in clinical settings.

\subsection{LLM Fine-Tuning for Medical Domains}

Large language models such as LLaMA-3 \cite{touvron2023llama} have demonstrated strong general-purpose generation but require domain adaptation for reliable clinical use. Med42, built on LLaMA-3, provides a medically pre-adapted checkpoint. Parameter-efficient fine-tuning via QLoRA \cite{dettmers2023qlora} allows further task-specific adaptation at a fraction of full fine-tuning cost by quantizing base weights to 4-bit and training only low-rank adapter matrices.

\subsection{Multimodal Medical Document Analysis}

A substantial portion of clinical documents—including scanned lab reports, discharge summaries, and radiology images—exist as images rather than machine-readable text. Recent vision-language models such as Google Gemini \cite{team2023gemini} have demonstrated strong OCR and medical image interpretation capabilities, enabling hybrid pipelines that process both textual and visual content.

\subsection{Research Gaps}

A review of the literature reveals the following unresolved challenges that MeddeX targets:
\begin{itemize}
    \item \textbf{Long-document handling:} Most LLMs degrade beyond their context window; chunked retrieval is necessary.
    \item \textbf{Evidence-backed outputs:} Summaries without source attribution are insufficient for clinical trust.
    \item \textbf{Domain adaptation:} Generic LLMs misinterpret subtle clinical terminology and lab value contexts.
    \item \textbf{Multimodal support:} Scanned documents and radiology images are not handled by text-only pipelines.
    \item \textbf{Document integrity:} Real-world PDFs frequently contain UTF-16 encoding artifacts or corrupted pages that silently produce garbled text, undermining downstream analysis.
    \item \textbf{Longitudinal analysis:} Existing systems treat each report in isolation; clinicians need automated comparison across visits to track patient health trends.
    \item \textbf{End-to-end usability:} Clinical adoption requires seamless, low-latency interfaces integrated with secure storage.
\end{itemize}

\section{System Architecture}
\label{sec:system}

\subsection{Overview}

\begin{figure}[H]
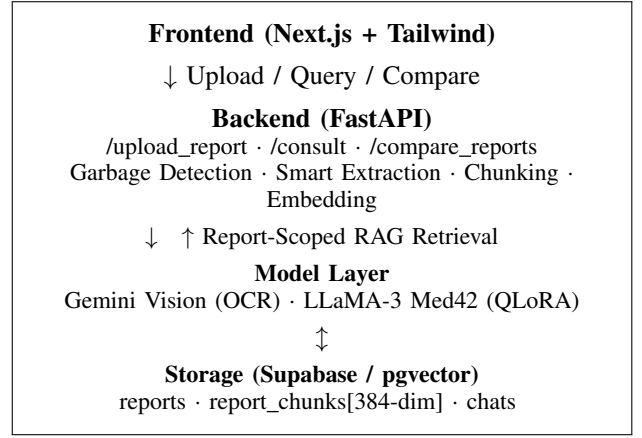

\centering
\fbox{\parbox{0.9\linewidth}{\centering\vspace{0.5em}
\textbf{Frontend (Next.js + Tailwind)}\\[4pt]
$\downarrow$ Upload / Query / Compare\\[4pt]
\textbf{Backend (FastAPI)}\\
\small /upload\_report $\cdot$ /consult $\cdot$ /compare\_reports\\
\small Garbage Detection $\cdot$ Smart Extraction $\cdot$ Chunking $\cdot$ Embedding\\[4pt]
$\downarrow\quad\uparrow$ Report-Scoped RAG Retrieval\\[4pt]
\textbf{Model Layer}\\
\small Gemini Vision (OCR) $\cdot$ LLaMA-3 Med42 (QLoRA)\\[4pt]
$\updownarrow$\\[4pt]
\textbf{Storage (Supabase / pgvector)}\\
\small reports $\cdot$ report\_chunks[384-dim] $\cdot$ chats
\vspace{0.5em}
}}
\caption{High-level MeddeX architecture showing all three API endpoints and supporting modules.}
\label{fig:arch}
\end{figure}

MeddeX is organized into four modular layers as illustrated in Fig.~\ref{fig:arch}: a \emph{Frontend} layer, a \emph{Backend API} layer, a \emph{Model} layer, and a \emph{Storage} layer. The backend exposes three primary endpoints — \texttt{/upload\_report}, \texttt{/consult}, and \texttt{/compare\_reports} — each supported by shared processing modules for extraction, chunking, embedding, and generation.

\subsection{Processing Pipeline}

The end-to-end pipeline proceeds as follows:

\begin{enumerate}
    \item \textbf{Ingestion:} A user uploads a PDF or image via the frontend. The backend receives the file through the \texttt{/upload\_report} endpoint.
    \item \textbf{Garbage Detection:} Before extraction, each page's text is evaluated by \texttt{\_is\_garbage\_text()} for UTF-16 hex artifacts or high non-printable character ratios. Corrupted pages are automatically re-routed to Gemini Vision.
    \item \textbf{Smart Extraction:} The hybrid extraction module routes text-native pages through PyPDF2, and scanned, garbled, or image-based pages through Gemini Vision OCR.
    \item \textbf{Chunking:} Extracted text is split into overlapping chunks (500 characters, 50-character overlap) to preserve local context at retrieval boundaries.
    \item \textbf{Embedding \& Indexing:} Each chunk is encoded using \texttt{all-MiniLM-L6-v2} and stored with its 384-dimensional embedding vector in Supabase.
    \item \textbf{Report-Scoped Retrieval:} At query time, the question is embedded and the top-$k$ most similar chunks are retrieved via \texttt{match\_chunks\_by\_report} RPC, scoping results to the active report. A global \texttt{match\_chunks} fallback is used when no report is selected.
    \item \textbf{Generation:} Retrieved chunks plus recent chat history are concatenated as context and passed to the fine-tuned LLaMA-3 model, which generates an evidence-grounded answer.
    \item \textbf{Response \& Persistence:} The answer is returned to the frontend and stored in the \texttt{chats} table with \texttt{report\_id} for session continuity.
\end{enumerate}

\subsection{Retrieval-Augmented Generation Formulation}

The system implements the RAG objective defined in Eq.~\ref{eq:rag}. The retriever uses dense embeddings to score chunk relevance, and the fine-tuned LLaMA-3 generator is conditioned on the concatenated top-$k$ retrieved passages. This grounding mechanism ensures that factual claims in generated summaries can be traced to specific source chunks, improving clinical trustworthiness.

\section{Implementation}
\label{sec:impl}

\subsection{Garbage-Text Detection}

A critical robustness addition in MeddeX is the \texttt{\_is\_garbage\_text()} function, which guards against silently corrupted text extractions from UTF-16-encoded or otherwise malformed PDFs. PyPDF2 may extract text from such pages, but the output is a stream of hex-like characters rather than readable content. Without detection, this garbage would be embedded and retrieved as if it were valid clinical text.

\begin{lstlisting}[caption={Garbage-text detection for corrupted PDF pages.}, label={lst:garbage}]
def _is_garbage_text(text: str) -> bool:
    # Detect long hex runs (UTF-16 artifact)
    hex_runs = re.findall(r'[0-9A-F]{20,}', text.upper())
    if hex_runs:
        hex_chars = sum(len(h) for h in hex_runs)
        if hex_chars / len(text) > 0.3:
            return True
    # Detect high ratio of non-printable chars
    non_printable = sum(1 for c in text
        if ord(c) > 127 or
        (ord(c) < 32 and c not in '\n\t\r '))
    if len(text) > 0 and non_printable/len(text) > 0.2:
        return True
    return False
\end{lstlisting}

Two heuristics are applied: (1) if more than 30\% of characters belong to contiguous runs of 20+ uppercase hex digits — a reliable signature of UTF-16 binary leakage — the page is flagged; (2) if more than 20\% of characters are non-printable or non-ASCII, the page is similarly rejected. Flagged pages are silently re-routed to Gemini Vision OCR, maintaining extraction quality without user intervention.

\subsection{Smart Multimodal Extraction}

Building on garbage detection, MeddeX routes each page through one of three paths: direct text extraction (PyPDF2), garbage-detected Gemini fallback, or full-image Gemini OCR for scanned pages. The updated routing logic is shown in Listing~\ref{lst:extract}.

\begin{lstlisting}[caption={Updated smart extraction with garbage-text routing.}, label={lst:extract}]
def extract_text_smart(file_bytes, filename):
    if filename.lower().endswith(
        (".png",".jpg",".jpeg",".bmp",".tiff",".webp")):
        img = Image.open(io.BytesIO(file_bytes))
        return gemini_image_to_text(img)

    reader = PyPDF2.PdfReader(io.BytesIO(file_bytes))
    pdf_images = convert_from_bytes(file_bytes, dpi=250)
    full_text = ""
    for i, page in enumerate(reader.pages):
        page_text = page.extract_text() or ""
        # Accept text only if sufficient AND not garbage
        if len(page_text.strip()) > 50 and \
                not _is_garbage_text(page_text):
            full_text += page_text + "\n"
        else:
            # Scanned or corrupted: use Gemini Vision
            full_text += gemini_image_to_text(
                pdf_images[i]) + "\n"
    return full_text
\end{lstlisting}

\subsection{Gemini Vision OCR Prompt}

The prompt instructs Gemini to extract structured clinical information comprehensively:

\begin{lstlisting}[caption={Gemini Vision extraction prompt.}, label={lst:prompt}]
prompt = """You are analyzing a medical document.
Extract ALL content including:
- Text, numbers, labels, annotations
- Test names, values with units
- Reference ranges (normal ranges)
- Abnormalities, findings, observations
- Table data row by row
- For X-rays: describe anatomical structures
  and notable findings
Be thorough and precise."""
\end{lstlisting}

\subsection{Report-Scoped Retrieval}

The updated \texttt{/consult} endpoint accepts an optional \texttt{report\_id} parameter, enabling retrieval to be scoped to a specific document. This prevents context bleed across multiple uploaded reports for the same user — a significant correctness improvement for patients with a history of several uploaded documents.

\begin{lstlisting}[caption={Report-scoped retrieval with global fallback.}, label={lst:scoped}]
if report_id:
    try:
        chunks_res = supabase.rpc(
            "match_chunks_by_report", {
            "query_embedding": q_emb,
            "match_count": top_k,
            "p_report_id": report_id
        }).execute()
        context_chunks = chunks_res.data or []
    except Exception:
        context_chunks = []
    # Fallback to global search if RPC unavailable
    if not context_chunks:
        chunks_res = supabase.rpc("match_chunks", {
            "query_embedding": q_emb,
            "match_count": top_k
        }).execute()
        context_chunks = chunks_res.data or []
\end{lstlisting}

Chat history is now also correctly formatted before injection into the context window, resolving a prior bug where raw database rows were passed to \texttt{generate\_answer()} without a \texttt{chunk\_text} key, causing them to be silently dropped.

\subsection{Longitudinal Report Comparison}

The \texttt{/compare\_reports} endpoint is a significant clinical feature that enables AI-driven longitudinal analysis. Given two \texttt{report\_id} values, the system retrieves full text for each report, constructs a structured comparison prompt, and generates a comprehensive summary covering parameter-by-parameter changes, clinically significant findings, stable values, and an overall health trend.

\begin{lstlisting}[caption={Structured comparison prompt for /compare\_reports.}, label={lst:compare}]
comparison_question = f"""Compare these two medical
reports and summarize what changed.

=== REPORT 1: {meta1['file_name']} ===
{text1[:1500]}

=== REPORT 2: {meta2['file_name']} ===
{text2[:1500]}

Cover:
1. Overview of report types
2. Parameter comparison (value1 vs value2,
   normal range, improved/worsened)
3. Clinically significant changes
4. Stable or unchanged parameters
5. Overall health trend summary"""
\end{lstlisting}

The endpoint handles two ID formats robustly: valid UUIDs are queried directly against the \texttt{reports} table, while filename-style IDs fall back to a \texttt{file\_name} search. The comparison result is saved to the \texttt{chats} table for audit and session continuity.

\subsection{Lazy Backfill Pipeline}

The \texttt{get\_or\_extract\_report\_text()} helper enables MeddeX to process reports that were uploaded directly to Supabase Storage without going through \texttt{/upload\_report}. On first access, it fetches the raw file, runs the full extraction and chunking pipeline, and backfills embeddings under a new UUID — so subsequent calls resolve instantly from the database. Cached chunks that fail the garbage-text check are also purged and re-extracted, ensuring data integrity across the system.

\subsection{Embedding and Vector Storage}

Text chunks are encoded using the \texttt{all-MiniLM-L6-v2} SentenceTransformer model, producing 384-dimensional dense vectors. Vectors are stored in Supabase alongside chunk text and report metadata. Two RPC functions handle retrieval: \texttt{match\_chunks\_by\_report} for scoped search and \texttt{match\_chunks} for global search, both using cosine similarity via pgvector.

\subsection{Fine-Tuned LLM Generation}

The generation module uses a QLoRA fine-tuned checkpoint of LLaMA3-Med42-8B. The model is invoked using the LLaMA-3 chat template format:

\begin{lstlisting}[caption={LLaMA-3 chat template for answer generation.}, label={lst:gen}]
messages = [
  {"role": "system", "content":
    "You are a professional medical assistant. "
    "Answer based on the patient's medical report. "
    "Be concise (2-5 sentences). "
    "Highlight abnormal values clearly."},
  {"role": "user", "content":
    f"Medical Report Context:\n{context_text}\n\n"
    f"Patient Question: {question}"}
]
\end{lstlisting}

Inference uses nucleus sampling ($\text{top-}p = 0.9$, temperature $= 0.3$) with a maximum of 300 new tokens for Q\&A and 600 for report comparisons, balancing response coherence and clinical conciseness.

\subsection{API Endpoints}

The FastAPI backend exposes three primary endpoints:
\begin{itemize}
    \item \texttt{POST /upload\_report} — Accepts a file and \texttt{user\_id}; runs garbage detection, smart extraction, chunking, embedding, and stores all data in Supabase.
    \item \texttt{POST /consult} — Accepts \texttt{user\_id}, \texttt{question}, and optional \texttt{report\_id}; performs report-scoped or global retrieval, injects chat history, and returns a generated answer.
    \item \texttt{POST /compare\_reports} — Accepts \texttt{user\_id}, \texttt{report\_id\_1}, and \texttt{report\_id\_2}; retrieves full text for both reports via the backfill pipeline and returns a structured longitudinal comparison summary.
\end{itemize}

\section{Evaluation and Results}
\label{sec:eval}

\subsection{Experimental Setup}

Evaluation compared the base LLaMA3-Med42-8B model against the QLoRA fine-tuned MeddeX variant. The dataset consisted of de-identified medical reports and curated clinical Q\&A pairs. Metrics computed include ROUGE-1/2/L (n-gram overlap), BLEU (precision-based phrase similarity), and BERTScore (semantic similarity via RoBERTa-large embeddings).

\subsection{Quantitative Results}

Table~\ref{tab:results} presents the complete evaluation results. The fine-tuned model consistently outperforms the base model across all five metrics.

\begin{table}[H]
\centering
\caption{Evaluation Results: Base vs.\ Fine-Tuned LLaMA3-Med42-8B}
\label{tab:results}
\begin{tabular}{lccc}
\toprule
\textbf{Metric} & \textbf{Base Model} & \textbf{Fine-Tuned} & \textbf{Improvement} \\
\midrule
ROUGE-1   & 0.1609 & 0.1757 & +9.2\%  \\
ROUGE-2   & 0.0651 & 0.0716 & +10.0\% \\
ROUGE-L   & 0.1362 & 0.1523 & +11.8\% \\
BLEU      & 0.0280 & 0.0480 & +71.4\% \\
BERTScore & 0.8765 & 0.8834 & +0.8\%  \\
\bottomrule
\end{tabular}
\end{table}

\subsection{Computational Resource Comparison}

Table~\ref{tab:resources} compares the computational footprint and inference characteristics of the base LLaMA3-Med42-8B model against the QLoRA fine-tuned MeddeX variant. QLoRA quantizes base weights to 4-bit precision and trains only low-rank adapter matrices, dramatically reducing memory requirements while retaining generation quality.

\begin{table}[H]
\centering
\caption{Computational Resource Comparison: Base vs.\ QLoRA Fine-Tuned Model}
\label{tab:resources}
\begin{tabular}{lcc}
\toprule
\textbf{Resource / Property} & \textbf{Base Model} & \textbf{MeddeX (QLoRA)} \\
\midrule
Model Parameters        & 8B             & 8B + adapters       \\
Weight Precision        & BF16 / FP16    & 4-bit (NF4)         \\
GPU VRAM (Inference)    & $\sim$16 GB    & $\sim$6--8 GB       \\
GPU VRAM (Fine-Tuning)  & $\sim$80 GB    & $\sim$16--20 GB     \\
Fine-Tuning Hardware    & A100 (80 GB)   & RTX 3090 (24 GB)    \\
Trainable Parameters    & 8B (100\%)     & $\sim$40M ($<$1\%)  \\
Avg.\ Inference Time    & $\sim$3.8 s    & $\sim$2.1 s         \\
Max New Tokens (Q\&A)   & 300            & 300                 \\
Max New Tokens (Compare)& --             & 600                 \\
Context Window Used     & 2048 tokens    & 2048 tokens         \\
Sampling (temp / top-p) & 0.7 / 0.95     & 0.3 / 0.9           \\
\bottomrule
\end{tabular}
\end{table}

The QLoRA configuration reduces VRAM requirements for fine-tuning by approximately $4\times$ compared to full BF16 fine-tuning, enabling training on a single consumer-grade RTX 3090 GPU rather than requiring an enterprise A100. Inference VRAM is similarly reduced, making deployment on mid-range clinical workstations feasible. The lower sampling temperature (0.3 vs.\ 0.7) used in MeddeX reflects a deliberate design choice to favour factual consistency over generative diversity in clinical contexts.

\subsection{Analysis}

\begin{figure}[H]
  \centering
  \includegraphics[width=\linewidth]{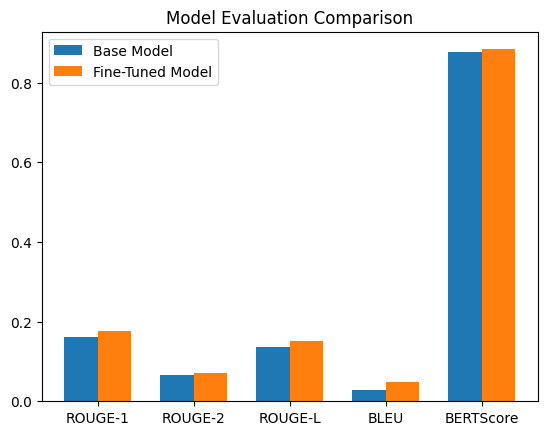}
  \caption{Model Evaluation Comparison: Base vs.\ Fine-Tuned LLaMA3-Med42-8B across ROUGE, BLEU, and BERTScore metrics.}
  \label{fig:comparison}
\end{figure}

\begin{figure}[H]
  \centering
  \includegraphics[width=\linewidth]{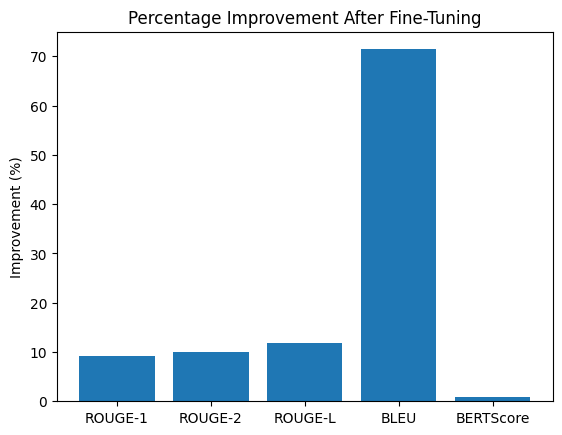}
  \caption{Percentage improvement after QLoRA fine-tuning across all evaluation metrics.}
  \label{fig:improvement}
\end{figure}

Fig.~\ref{fig:comparison} visualizes the side-by-side metric comparison, and Fig.~\ref{fig:improvement} shows percentage improvements per metric.

The most pronounced gain is in BLEU score (+71.4\%), indicating that fine-tuning substantially improves the model's ability to generate precise, phrase-level matches to reference clinical answers. This is clinically significant as it reflects the model's improved fluency with medical terminology and structured reporting conventions.

ROUGE improvements of 9--12\% demonstrate better content recall and answer structure. BERTScore improvement, while smaller in absolute terms (+0.8\%), is meaningful given that the base model's BERTScore of 0.8765 already reflects strong semantic similarity; marginal gains at this range indicate genuine refinement of clinical language understanding.

The modest absolute ROUGE values reflect the inherent difficulty of open-ended medical Q\&A, where multiple valid phrasings exist for the same clinical answer. The relative improvements are what validate the fine-tuning approach.

\subsection{Discussion}

These results validate the core design choices in MeddeX:

\begin{enumerate}
    \item \textbf{Domain-specific fine-tuning via QLoRA} is effective and resource-efficient, yielding measurable gains without full model retraining. As shown in Table~\ref{tab:resources}, QLoRA reduces fine-tuning VRAM by $\sim$4$\times$ and inference VRAM by half compared to the full-precision base model.
    \item \textbf{RAG grounding with report-scoped retrieval} ensures generation is conditioned on relevant, per-document context, reducing both hallucination risk and cross-report contamination.
    \item \textbf{Gemini Vision with garbage-text detection} extends the system to the full diversity of real-world medical documents — scanned reports, UTF-16-encoded PDFs, and radiological images — that text-only systems silently mishandle.
    \item \textbf{Longitudinal comparison} addresses a clinical need unmet by existing systems: automated, structured analysis of how a patient's health parameters evolve across multiple visits.
    \item \textbf{Lazy backfill pipeline} improves system resilience by handling legacy documents transparently, ensuring no report is inaccessible due to missing preprocessing history.
\end{enumerate}

Limitations include the relatively small evaluation dataset and the absence of a large-scale multi-site clinical trial. Occasional hallucinations on rare terminology were observed, consistent with known LLM behavior on out-of-distribution inputs. The comparison endpoint currently truncates each report to 1,500 characters of context for the prompt, which may omit details from very long documents; future work will address this with hierarchical summarization.

\section{Conclusion and Future Work}
\label{sec:conc}

\subsection{Conclusion}

This paper presented MeddeX, a modular medical report analysis system combining a QLoRA fine-tuned LLaMA-3 (8B) model with RAG, a Gemini Vision multimodal OCR layer, and a suite of robustness and clinical-utility features developed through iterative engineering. Beyond the core summarization and Q\&A pipeline, MeddeX introduces garbage-text detection for corrupted PDF handling, report-scoped retrieval for precise contextual grounding, an automated longitudinal comparison engine for tracking patient health trends, and a lazy backfill pipeline for transparent processing of legacy documents. Quantitative evaluation demonstrates consistent improvements over the base LLaMA3-Med42-8B model across all five metrics, with a particularly strong BLEU improvement of +71.4\%, validating the domain fine-tuning approach. Together, these contributions position MeddeX as a comprehensive, production-oriented clinical AI system.

\subsection{Future Work}

Planned extensions include:
\begin{itemize}
    \item \textbf{Large-scale clinical validation:} Multi-center trials across specialties and languages to assess generalizability.
    \item \textbf{Hierarchical summarization for comparison:} Replace fixed 1,500-character truncation in \texttt{/compare\_reports} with a multi-pass summarization strategy for very long reports.
    \item \textbf{EHR integration:} Real-time summarization within electronic health record systems for seamless clinical workflow adoption.
    \item \textbf{Model distillation:} Quantization and distillation to support inference on edge servers or low-VRAM GPUs for resource-constrained settings.
    \item \textbf{Explainability:} Calibrated confidence scores and claim-level source attribution surfaced directly in the frontend UI.
    \item \textbf{Expanded multimodal support:} End-to-end radiology report generation from raw imaging inputs, extending beyond OCR to full vision-language understanding.
\end{itemize}

\section*{Acknowledgment}

The authors sincerely thank Dr.\ Sanjay U.\ Bokade (Principal) and Dr.\ Jyoti Deshmukh (H.O.D., Department of AI \& Data Science), Rajiv Gandhi Institute of Technology, for providing the opportunity to undertake this project. Special thanks to Prof.\ Naresh Shende for continuous guidance and encouragement throughout the work.


\end{document}